\documentclass[10pt,twocolumn,letterpaper]{article}

\usepackage[pagenumbers]{iccv} 

\usepackage{algorithm}
\usepackage{algpseudocode}
\usepackage{makecell}
\usepackage{multirow}
\usepackage[accsupp]{axessibility}
\algnewcommand{\To}{\textbf{To }}
\algnewcommand\Input{\item[\textbf{Input:}]}%
\algnewcommand\Output{\item[\textbf{Output:}]}%

\definecolor{iccvblue}{rgb}{0.21,0.49,0.74}
\usepackage[pagebackref,breaklinks,colorlinks,allcolors=iccvblue]{hyperref}

\def\paperID{BIC-8} 
\def\confName{ICCV}
\def\confYear{2025}

\title{Preserving instance continuity and length in segmentation through connectivity-aware loss computation}

\author{
Karol Szustakowski\textsuperscript{1}, Luk Frank\textsuperscript{2}, Julia Esser\textsuperscript{2}, Jan Gründemann\textsuperscript{2,3}, Marie Piraud\textsuperscript{1}\\
\textsuperscript{1} Helmholtz AI, Helmholtz Munich, Neuherberg, Germany \\
\textsuperscript{2} German Center for Neurodegenerative Diseases (DZNE), Bonn, Germany \\ 
\textsuperscript{3} University of Bonn, Medical Faculty, Bonn, Germany \\
{\tt\small \{karol.szustakowski,marie.piraud\}@helmholtz-munich.de} \\
{\tt\small \{luk.frank,julia.esser,jan.grundemann\}@dzne.de} \\
}

\begin{document}
\maketitle
\thispagestyle{empty}
\textit{
In many biomedical segmentation tasks, the preservation of elongated structure continuity and length is more important than voxel-wise accuracy. We propose two novel loss functions, Negative Centerline Loss and Simplified Topology Loss, that, applied to Convolutional Neural Networks (CNNs), help preserve connectivity of output instances. Moreover, we discuss characteristics of experiment design, such as downscaling and spacing correction, that help obtain continuous segmentation masks. We evaluate our approach on a 3D light-sheet fluorescence microscopy dataset of axon initial segments (AIS), a task prone to discontinuity due to signal dropout.
Compared to standard CNNs and existing topology-aware losses, our methods reduce the number of segmentation discontinuities per instance, particularly in regions with missing input signal, resulting in improved instance length calculation in downstream applications. Our findings demonstrate that structural priors embedded in the loss design can significantly enhance the reliability of segmentation for biological applications.
}
\section{Introduction}
\label{section:introduction}
\par Segmentation, as a supervised objective, often serves as a proxy for multiple downstream processing applications. In medical settings, researchers often prefer to annotate data with segmentation masks due to the abundance of annotation software packages and models tailored specifically for the segmentation target.
Biomedical imaging applications, such as cell instance segmentation, vessel segmentation, organ segmentation, and similar, often utilize those segmentations to compute morphological features of the structures of interest. This leads to a misalignment problem between the features important for a given downstream task, \eg vessel width calculation, and the optimization target of the segmentation model, usually the pixel-wise binary cross-entropy \cite{DBLP:journals/corr/abs-1911-01685, MA2021102035, herrera2022impactlossfunctiondeep}. The former is generally computed algorithmically, by post-processing the segmentation mask, while the latter is optimized directly, through gradient minimization of a specified loss function. 
This misalignment is not a problem for technical reporting, which is often concerned with standard metrics such as accuracy or Dice, but becomes problematic in the applicative scenario \cite{Maier_Hein_2024}. Continuing the previous vessel example, although a model may achieve highly accurate global segmentation, this does not necessarily translate to precise vessel width measurements derived from the segmentation mask~\cite{khanal2019dynamicdeepnetworksretinal}.

\subsection{Motivation}
The axon initial segment (AIS) is a highly organized microdomain located at the proximal part of the axon. It plays a central role in neuronal function as the site of action potential initiation, maintaining neuronal polarity, and regulating axonal transport \cite{grubb_burrone_210, rasband2010, YuLi:e768}. Beyond its structural role, the AIS has emerged as a site of functional plasticity, capable of adapting its length, position, and molecular composition in response to physiological and pathological changes \cite{kole_2008, kuba_2010, yoshimura_2014, kole_2018}. Such modifications directly influence neuronal excitability, highlighting the AIS as a dynamic and responsive structure within the neuron. Understanding AIS morphology under different conditions is therefore essential for uncovering how neurons adjust their signalling properties and maintain network stability \cite{grundemann_2010, jamann_2021, garrido_2023, FREAL2025649}.

Light-sheet fluorescence expansion microscopy (LSFEM) integrates the tissue-clearing advantage of expansion microscopy (ExM) with the fast volumetric imaging capabilities of light-sheet fluorescence microscopy (LSFM), resulting in a powerful tool for rapid, high-throughput acquisition of large tissue volumes \cite{doi:10.1126/science.1260088, burgers_2019}. This combination makes it particularly well suited for applications requiring broad spatial coverage without compromising structural detail. Notably, LSFEM achieves approximately 17-fold faster imaging than conventional high-resolution confocal microscopy when scanning cubic millimeter-scale brain regions \cite{burgers_2019}. This makes it exceptionally efficient for large-scale studies to investigate AIS morphology. Through enabling rapid and comprehensive 3D visualization of AIS across extended brain areas, LSFEM supports high-throughput structural analysis beyond the limits of standard imaging techniques \cite{tillberg_2016, burgers_2019}.
For downstream research questions, an important feature of an AIS is its length, start and end point, which is often determined by a human by tracing the AIS semi-automatically. This human-in-the-loop tracing proves impractical for thousands of axons, particularly in 3D. 

Multiple solutions currently exist for AIS and axon tracing, which are often adapted to the acquisition modality. For instance, using the APP2 algorithm \cite{app2}, reconstructions of neurons can be obtained from fluorescence images and subjected to further analysis \cite{XU2024109264}. Neuron tracing can also be performed using software packages such as \textit{NeuroTreeTracer} \cite{NeuroTreeTracer}, which can be used by importing $z$-stack fluorescent images into Fiji/ImageJ \cite{Schindelin2012}, merging channels, and generating sum projections. The resulting images are then processed using MATLAB scripts provided with the software to segment neurons, extract somas, trace neurites, and quantify fluorescent intensities along neurite paths with background correction \cite{DiRe2019AISImaging}. Tracing is often performed in a tool-assisted manner, using toolboxes such as \textit{AISuite} (\href{https://github.com/jhnnsrs/aisuite2}{github.com/jhnnsrs/aisuite2}), which allows measuring AIS properties by detecting and straightening axonal segments based on fluorescent markers \cite{Thome_2024}. The tool enables the extraction of intensity profiles, calculation of AIS distal points, and determination of physical dimensions using image metadata. In contrast to the aforementioned solutions, we aimed to develop an end-to-end solution that can be applied to light-sheet microscopy images without the need for human involvement beyond the labelling procedure.

\begin{figure}[htpb]
    \centering
    \begin{subfigure}{0.3\linewidth}
        \centering
        \includegraphics[height=0.8\linewidth,trim={0 0 4.7cm 0},clip,angle=90]{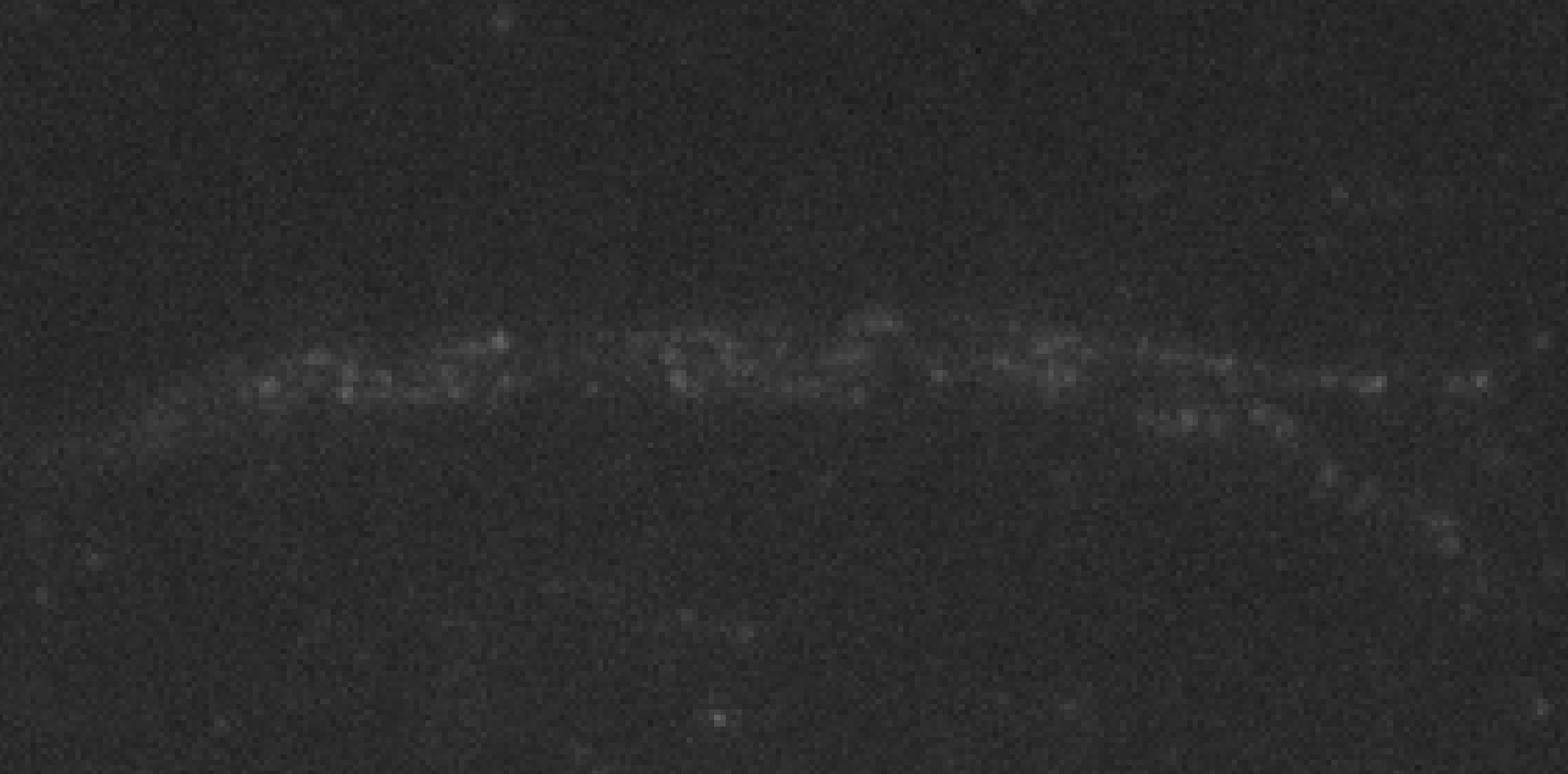}
        \caption{AIS image}
        \label{fig:disconnection:raw}
    \end{subfigure}
    \begin{subfigure}{0.3\linewidth}
        \centering
        \includegraphics[height=0.8\linewidth,trim={0 0 4.7cm 0},clip,angle=90]{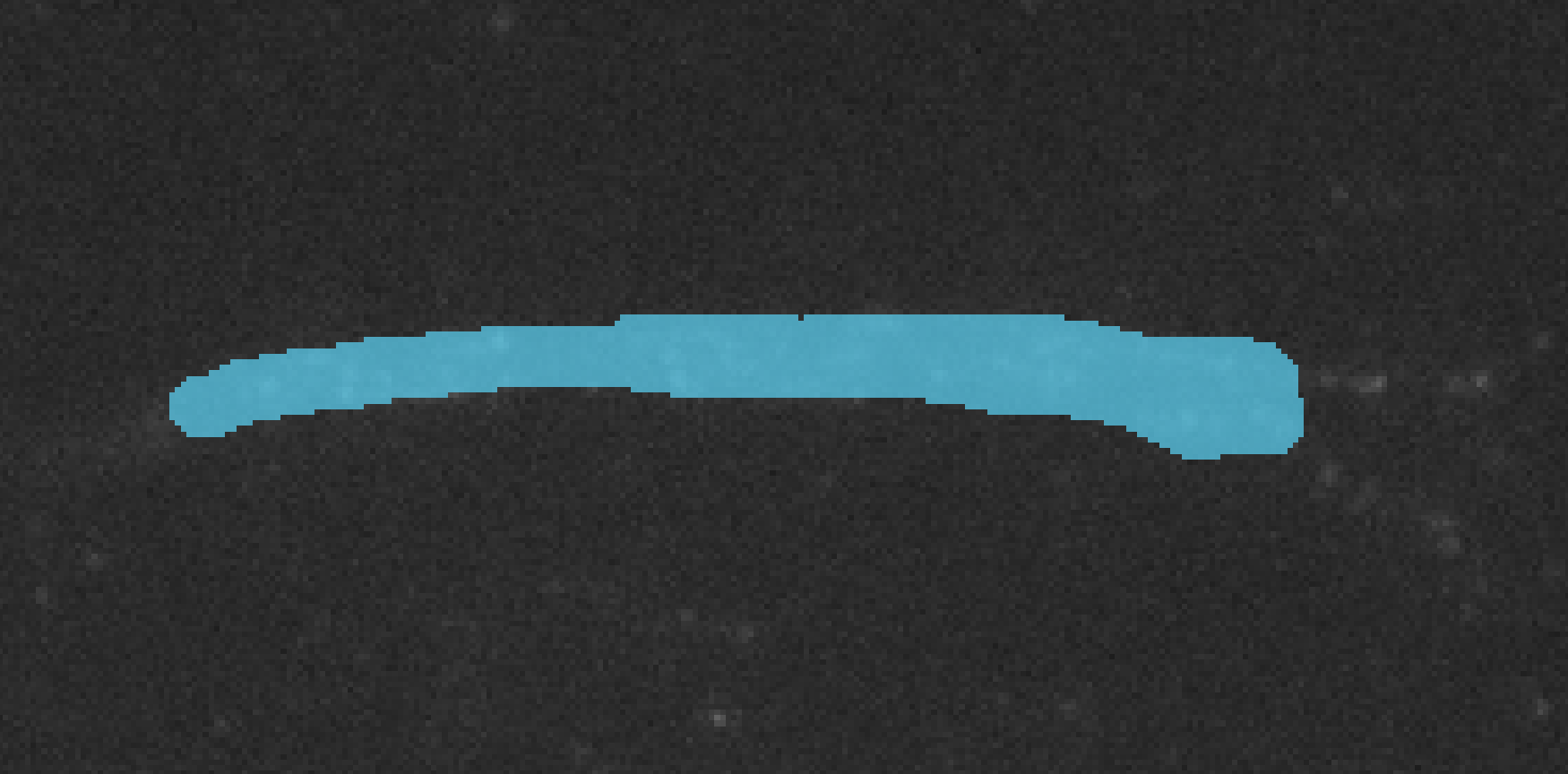}
        \caption{Label}
        \label{fig:disconnection:label}
    \end{subfigure}
    \begin{subfigure}{0.3\linewidth}
        \centering
        \includegraphics[height=0.8\linewidth,trim={0 0 4.7cm 0},clip,angle=90]{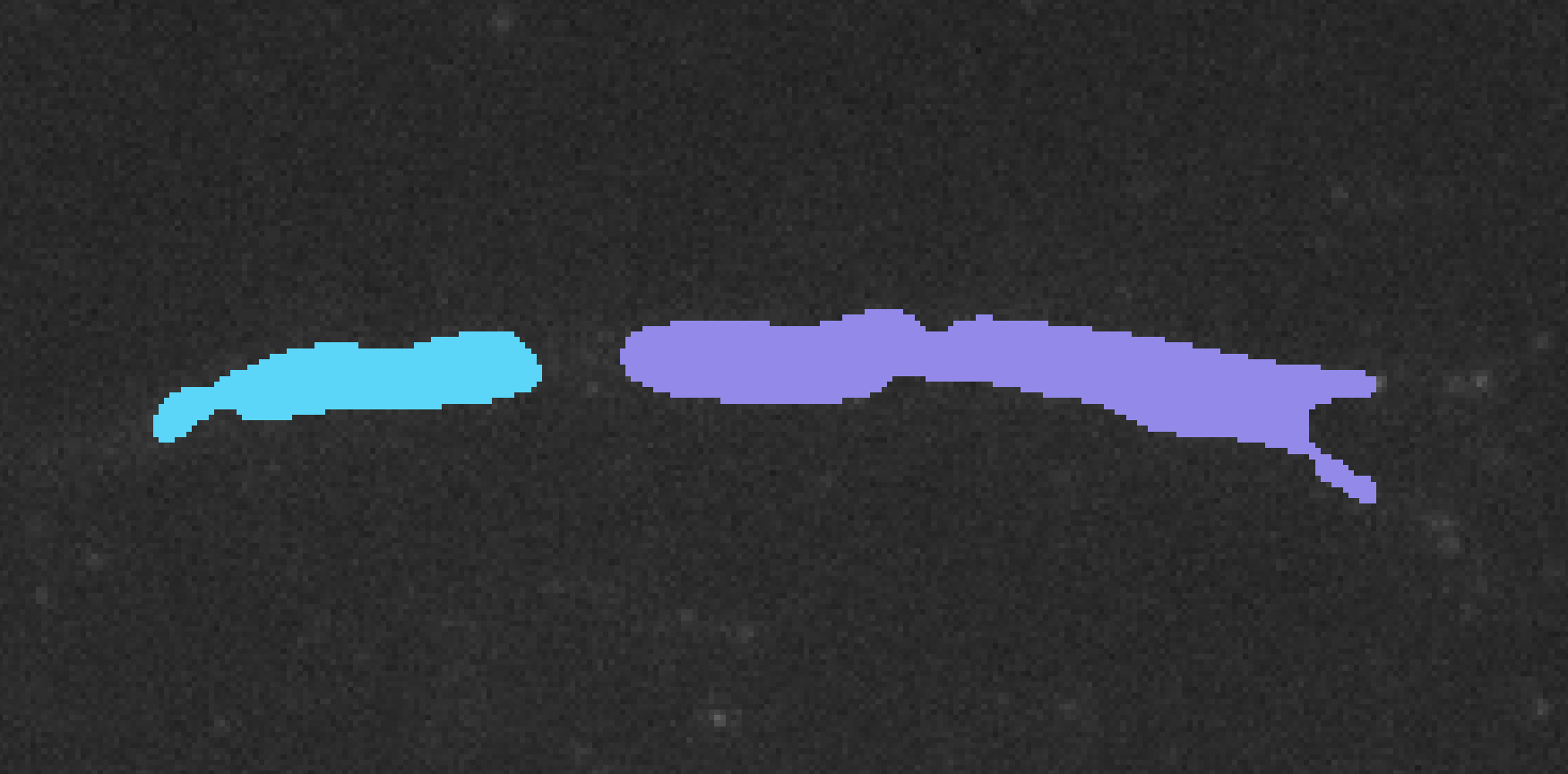}
        \caption{Model prediction}
        \label{fig:disconnection:pred}
    \end{subfigure}
    \caption{An example of model outputting disconnected labels due to input signal absence.}
    \label{fig:disconnection}
\end{figure}

We procured a 3D dataset of light-sheet fluorescence expansion microscopy images of AIS of mouse brains and their manually annotated segmentation masks, which were created using \textit{napari} \cite{napari2019}, to automate the length calculation task. AIS in medial prefrontal cortex were visualized with immunofluorescence against AnkG. As seen in Figure~\ref{fig:disconnection}, fluorescence microscopy of axonal structures often exhibits uneven signal strength and signal dropoff, which can be easily noticed and interpolated by humans, but can result in discontinuity of automatic predictions.  
This separation is detrimental for automatic AIS detection and length calculation, as every such flawed segmentation results in two or more detections of AIS instances with shorter length instead of one longer, correct AIS.
Motivated by this issue, we evaluate the available methods and propose two novel loss functions to resolve the discontinuity problem in segmentation, especially in scenarios of missing input signal.

\subsection{Related work}
Multiple methods exist that leverage the skeletonizations of the predictions, which, when compared with the skeletonizations of the labels, can be used to coerce the models to perform better if the skeleton is an important part of the prediction. This is true especially for elongated structures, since any discontinuity is also a disruption of the skeleton. Xu \etal propose \textit{CP-loss} \cite{cploss}, which penalizes skeleton misalignment based on the Euclidean distance of corresponding skeleton points. Similarly, Shit \etal propose \textit{clDice}, a topology-preserving loss \cite{cldice2021}, in which the skeleton of each object influences the loss equally, due to a scaling term dependent on the skeleton's size.
Another common approach is to optimize the alignment of topological descriptors, such as persistent Betti numbers \cite{persistence}. In imaging scenarios, persistent homology is usually built on top of cubical complexes \cite{cubdata}. Stucki \etal propose an efficient implementation of Betti Matching for topology-aware segmentation \cite{bettimatching}. On a similar note, Hu \etal propose a loss function that operates on the persistence diagrams to match the Betti numbers \cite{diagramtopo}. An overarching downside of most methods based on persistent homology is the computational complexity of such methods; commonly, the time required to compute the topological descriptors of which the loss is composed constitutes the majority of the training time.
In contrast to approaches based on direct minimization of a differentiable function, methods such as \textit{Topograph}~\cite{topograph} abandon the notion of implementing a differentiable method and processing the predicted masks separately, reinforcing loss in regions that alter the topology of the predictions. Due to not adhering to differentiability, Lux \etal can implement more advanced graph-based logic.

In this work, we take inspiration from both \textit{Topograph} and \textit{clDice}. We propose two novel loss functions, \textit{Negative Centerline Loss}, a skeleton-based approach that focuses on connectivity, and \textit{Simplified Topology Loss}, which coerces the model to improve performance on regions that cause discontinuities. 

\section{Methods}
\begin{figure*}[htpb]
    \centering
    \begin{subfigure}{\textwidth}
        \centering
        \includegraphics[width=\textwidth]{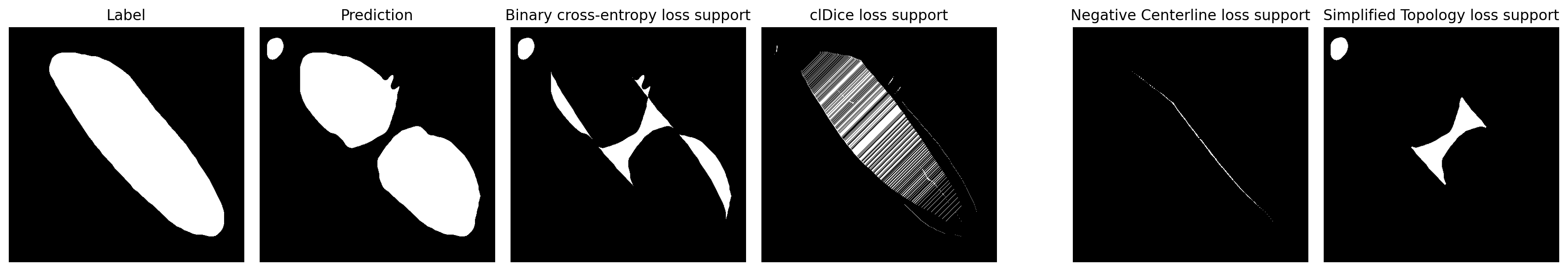}
        \caption{Loss support comparison for a simple discontinuity}
        \label{fig:support_comp:disconnectivity}
    \end{subfigure}
    \begin{subfigure}{\textwidth}
        \centering
        \includegraphics[width=\textwidth]{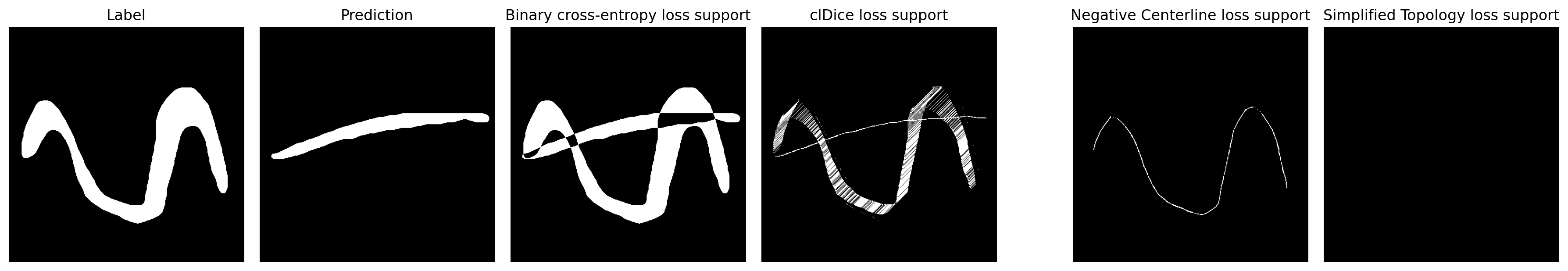}
        \caption{Loss support comparison for bending}
        \label{fig:support_comp:bend}
    \end{subfigure}
    \caption{Comparison of the supports of selected loss functions. The striped pattern of \textit{clDice} loss support is a result of sequential application of the pooling operations, firstly in the $y$, and subsequently in the $x$ direction. Merging those two passes into a single pooling operation eliminates those artifacts, but produces a centerline with spurs. In this work, we adhere to the \textit{clDice} authors' two-pass implementation~\cite{cldice_github}. Remarkably, the \textit{Simplified Topology Loss} is zero for the bending example, as the label and prediction instances overlap, and both the label and the prediction are a single connected component.}
    \label{fig:support_comp}
\end{figure*}

We begin with a discussion of the drawbacks of current methods. Figure \ref{fig:support_comp} compares the supports of multiple loss functions. We define the support as the set of prediction pixels/voxels for which the gradient of the loss is non-zero. Figure \ref{fig:support_comp:disconnectivity} is an illustrative example of the most common discontinuity issue present when detecting elongated structures, with a noticeable split in the middle of the instance. The split is included in the support of binary cross-entropy, but so are multiple other regions that are not important for object length and connectivity. In \textit{clDice}, not only is the split included, but also the remaining parts of the centerline, along with multiple artifacts resulting from misalignment of the centerline. 
The issue is more noticeable in Figure \ref{fig:support_comp:bend}. Topologically, the two detections are the same, since they are both composed of a single connected component.  
Those observations help define a desirable continuity-preserving loss function:
\begin{itemize}
    \item the discontinuity should be reflected in the support of the loss function,
    \item misclassified pixels that do not cause discontinuity and constitute the outer borders of instances shall not have a big influence on the loss function. 
\end{itemize}
Both of the proposed loss functions are built with those properties in mind. Section \ref{sec:results} discusses the scenarios in which these loss functions excel and why we consider two proposals.

\subsection{Negative Centerline Loss}
\textit{Negative Centerline Loss}, introduced as Algorithm \ref{alg:negative_centerline}, measures the proportion of the label's centerline that does not intersect with the prediction. As seen in Figure \ref{fig:support_comp:disconnectivity}, the support of the loss function is the centerline of the label. More importantly, the influence of the whole centerline is minor, as it stems only from the division by the centerline length. The gradient's magnitude is the largest in the area of the discontinuity, as this is the part of the centerline that intersects with $1-P$, as seen in Algorithm \ref{alg:negative_centerline}. Looking at Figure \ref{fig:support_comp:bend}, we still see some excessive loss support in the case of bending, but in practice this case is a rare occurrence when segmenting elongated structures. The majority of the computation time of this method stems from the \textit{soft-skeleton} subroutine (Algorithm \ref{alg:soft_skeleton})  \cite{cldice2021}. Here, we perform the \textit{soft-skeleton} procedure until convergence (in contrast to \textit{clDice}, which uses a fixed number of iterations), therefore the worst-case time complexity is $O(nd)$, where $n$ is the number of pixels (voxels in 3D), and $d$ is the size of the largest dimension of the image. Realistically, for tubular objects segmentation, the time complexity is $O(nw)$, where $w$ is the radius of the largest tubular object. Similarly to \textit{clDice} the number of iterations can be limited, but in practice the computation time of the \textit{Negative Centerline Loss} is rarely noticeable. 

\begin{algorithm}
\caption{\textit{Negative Centerline Loss}\\
$A \circ B$ denotes the Hadamard product of $A$ and $B$\\
$|A|$ denotes the sum of all elements of $A$}\label{alg:negative_centerline}
\begin{algorithmic}
\Input{Label $L$, prediction $P$}
\State $L_{CL} \gets \text{soft-skeleton}(L)$
\State $NI_{CL} \gets (\textbf{1}-P) \circ L_{CL}$
\State $NegativeCenterline \gets {|NI_{CL}| \over |L_{CL}|}$
\Output $NegativeCenterline$
\end{algorithmic}
\end{algorithm}

\begin{algorithm}
\caption{soft-skeleton}\label{alg:soft_skeleton}
\begin{algorithmic}
    \Input{Tensor $I$}
    \State $L_p \gets maxpool(minpool(I))$ 
    \State $CL \gets \text{ReLU}(I - L_p)$
    \State $I \gets minpool(I)$
    \While {$|I| > 0$}
        \State $I_{next} \gets minpool(I)$
        \State $L_p \gets maxpool(I_{next})$
        \State $CL \gets CL + \text{ReLU}((\textbf{1}-CL) \circ \text{ReLU}(I-L_p))$
        \State $I \gets I_{next}$
    \EndWhile
    \State $CL \gets minpool(maxpool(CL))$
    \Output{$CL$}
\end{algorithmic}
\end{algorithm}

\subsection{Simplified Topology Loss}
The \textit{Simplified Topology Loss} does not implement a differentiable loss function. Instead, similarly to \textit{Topograph}, we compute the regions important for preserving continuity, and reapply the binary cross-entropy loss within them.
The first step of computing the \textit{Simplified Topology Loss} is thresholding the prediction to obtain a binary mask. This is reflected in Algorithm \ref{alg:find-regions}. We then analyse all the regions that are present in the label $L$, but are missing in the prediction $P$. Concretely, we are interested in regions of $\text{ReLU}(L-P)$, that border at least two regions of the prediction, as those are the regions that cause discontinuity of the instances. 

To analyse connectivity, it helps to dilute both the prediction and $\text{ReLU}(L-P)$ tensors. After connected-component labelling of these dilutions, a border is defined as a set of tensor indices such that both the component indexes of the diluted $\text{ReLU}(L-P)$ and the diluted prediction are non-zero.
\begin{algorithm}
\caption{find-regions \\
Binary dilations are performed with a hypercube structuring element with square connectivity equal to one.}
\label{alg:find-regions}
\begin{algorithmic}
    \Input{Tensors: label $L$, prediction $P$}
    \State $P \gets P > 0.5$    
    \State $P \gets \text{binary dilation of } P$
    \State $CC^P \gets \text{connected components labeling of }P$
    \State $D \gets \text{ReLU}(L-P)$
    \State $D \gets \text{binary dilation of } D$
    \State $CC^D \gets \text{connected components labeling of }D$
    \State $U \gets \{(d, p) : \exists_{\text{tensor index } I} \text{  } CC^P_I  = p \wedge CC^D_I = d\}$
    \State $U \gets \{(d, p) : d \ne 0 \wedge p \ne 0\}$
    \State $D_R \gets \left\{ d : \left| \left\{ p : (d, p) \in U \right\} \right| \ge 2 \right\}$
    \State $P_R \gets \{p : !\exists _d (d, p) \in U  \}$
    \State $R \gets \text{tensor defined for indices }I:\newline \hspace*{2em}  R_I = \begin{cases}
        0, & \text{if } CC^P_I \in R_P \text{ or } CC^D_I \in D_R \\
        1, & \text{otherwise}
    \end{cases}$
    \Output{R}
\end{algorithmic}
\end{algorithm}
Subsequently, having obtained all the pairs of region indices that overlap, we extract:
\begin{itemize}
    \item indices of the regions in $\text{ReLU}(L-P)$ that are connected to at least two distinct regions in the prediction,
    \item indices of the prediction regions that do not overlap with any region of the label.
\end{itemize}
Examples of both those region types can be seen in Figure~\ref{fig:support_comp:disconnectivity}. The hourglass-shaped region in the \textit{Simplified Topology Loss} support coincides with two distinct regions of the prediction. The additional small instance in the support is present because it does not overlap with any label instance, but is part of the prediction.
The result of the algorithm is a mask for which we recompute the binary cross-entropy loss.
The unique pairs $(d,p)$ of Algorithm \ref{alg:find-regions} can be computed in amortized linear time using hash tables. The same can be applied to the last three steps of the algorithm, that are composed of filtering by count and lookup. Therefore, the amortized time complexity of the \textit{find-regions} procedure is the same as of the connected components labelling procedure, which can be implemented in $O(n \alpha(n))$ time, where $n$ is the number of elements of the $L$ tensor (\eg the number of pixels of an image), and $\alpha$ is the inverse Ackermann function. 
We nominate the method as \textit{simplified}, as in contrast to most homology-based methods it only considers continuity, as we are not interested in analysing higher dimensional holes.

\subsection{Experimental setup}
We compare the results obtained for all the considered loss functions applied to training an \textit{nnU-Net}-generated U-Net architecture \cite{nnunet}. The U-Net network architecture \cite{unet} is a popular choice for 3D image segmentation in the microscopy setting, and, as backed by \textit{nnU-Net} authors Isensee \etal \cite{nnunet_revisited}, we did not notice major performance differences of the model itself when compared to transformer-based architectures.
The U-Net configuration was generated once, and shared between the loss functions. The model was always trained for 500 epochs, and the best validation checkpoint was later used for the predictions for each fold. All the results presented are obtained via a 5-fold cross-validation procedure. We report the mean values from those 5 folds for all the metrics.
In addition to \textit{clDice}, we also compare our results with a topology loss function. Topology loss is defined as the bottleneck distance between the persistence diagrams of the labels and the predictions, built upon cubical complexes. Although there are multiple ways to involve the bottleneck distance in the loss computation \cite{ozcelik2024topologyawarelossaortagreat, Zia_2024, Byrne_2023}, it is also possible to directly minimize it, enforcing direct topological matching.

We combine all evaluated loss functions with binary cross-entropy and Dice losses:
$$\mathcal{L} = w_{BCE} \mathcal{L}_{BCE} +  w_{Dice} \mathcal{L}_{Dice} +  w_{eval} \mathcal{L}_{eval}$$
$w_{BCE}$, $w_{Dice}$, and $w_{eval}$ are selected empirically to make the $w_{eval}$ as large as possible while still maintaining good segmentation quality improvement as measured by the Dice validation metric during the first epochs of training. Table \ref{tab:weights} lists the weights assigned to each evaluated loss function.

\begin{table}
\caption{Weights assigned for each evaluated loss function.}
\centering
    \label{tab:weights}
\begin{tabular}{|c||c|c|c|}
    \hline
    Run & $w_{BCE}$ & $w_{Dice}$ & $w_{eval}$ \\
    \hline
    \makecell{Baseline \textit{nnU-Net}} & 1 & 1 & N/A \\ 
    \hline
    clDice & 1 & 1 & 3 \\
    \hline 
    Topology loss & 1 & 1 & 1 \\
    \hline
    \makecell{Negative Centerline} & 1 & 1 & 3 \\
    \hline 
    \makecell{Simplified Topology} & 1 & 1 & $4 \over 5$\\
    \hline
\end{tabular}
\end{table}

\subsection{Performance measurement}
Standard methods for measuring segmentation performance are the \textit{Dice coefficient} or voxel accuracy. In the case of elongated object segmentation, we are interested in objects' lengths. Furthermore, a length of a single AIS is not as important as the distribution of lengths of AIS gathered from the whole microscopy volume. 
For this reason, we focus on metrics such as Wasserstein Distance of the length distributions, to compare the distributions of AIS lengths between the predictions and the labels, strictly standardized mean difference (SSMD) of means and standard deviations between the labels and predictions, precision, and recall. 
We define precision as the proportion of connected instances in the segmentation output that overlap with some instance from the input. Similarly, we define recall as the proportion of connected instances in the labels that overlap with some instance from the segmentation output. Importantly, once a prediction or label instance is matched with another prediction/label instance for precision or recall computation, those two instances cannot be matched again. For instance, if the segmentation procedure produces a correct but discontinuous AIS, one part of it will be matched with the label and contribute positively to both precision and recall, but the other part will not be matched. In this case the recall is $1.0$, since the label is matched with some prediction, but the precision is $0.5$. We note that some uncertainty is introduced by the labelling procedure, where AIS in dim or unclear areas of the volumes were missed by the annotators, but retrieved correctly by the model. Those instances contribute negatively towards precision.
Finally, we directly measure the average number of prediction instances each label instance overlaps with, if it overlaps with any (unmatched labels are not considered by this metric). Ideally, the value of this metric should be 1, indicating that each label instance overlaps with exactly one prediction instance.
\subsection{Length computation}
Throughout this text, we frequently refer to AIS length. We compute it by performing a skeletonization of the segmentation masks. Afterwards, we built a graph with nodes composed of skeleton junction and end points. The edges of the graph represent the distance between the graph points, which is the number of voxels between the junction/end points, with $z$-stack spacing considered. Finally, we find the diameter of the graph and consider it to be the AIS length. For the computation of performance metrics involving length, we discard any instances that intersect with the image border.  Other authors compute the AIS length by manually tracing the shortest path in the segmentation from the soma to the axon tip \cite{10.3389/fncel.2018.00146}, tracing the AIS by hand~\cite{10.3389/fncel.2016.00268}, or using a semi-automated approach, where straightened intensity profiles of ßIV-spectrin are analysed, and AIS is defined as the segment with intensity above 30\% of the maximum signal intensity \cite{10.3389/fncel.2017.00332}. Ultimately, to the best of our knowledge, there is no standardized way to compute AIS length on light-sheet microscopy images. In contrast to the aforementioned techniques, the method we propose is deterministic and does not require human intervention for computation. We remark that similar path-based solutions were previously employed, albeit outside the context of AIS~\cite{lee2024measureanythingrealtimemultistage, 6976882, shaperep}.
\section{Results}
\label{sec:results}
The dataset used for training is composed of 10 light-sheet microscopy volumes, gathered from 5 separate volume acquisitions, each performed on a different mouse, with two volumes per acquisition. We adapted a cross-validation approach, in which all volumes from a given acquisition formed the validation set, and all other volumes formed the training set. With that, the validation was always performed on previously unseen mouse and acquisition method combination. The volumes acquired are $2000\times2000$ voxels in size, with roughly 150 $z$-stacks per image. This resulted in volumes containing 16 to 56 AIS per volume.

We present the results obtained for multiple loss functions in Table \ref{tab:results}. All methods benefit greatly from downscaling the volumes before prediction.
\subsection{Predictions on downscaled volumes}
Before discussing the results obtained for multiple loss functions, we present a subsidiary result of this study, which is the effect of downscaling for preserving connectivity. 
The AIS in our dataset have varying lengths, with the mean ranging from 312 to 518 voxels, depending on the mouse.
This means that with the \textit{nnU-Net} autotuned patch size of $256\times256\times32$, AIS instances are either contained entirely within a single sliding window, or slightly larger. One could also expect a higher resolution of the input images to translate to better output segmentation quality.
Our results show the opposite. We notice very considerable performance gains when downscaling the $x$ and $y$ dimensions of the samples thrice. We train and evaluate the model on the downscaled samples, and then compute all the metrics of interest on the upsampled segmentation masks. According to our exploratory analysis, downscaling the $z$ stack was not beneficial. Similarly, we do not apply spacing correction, that is the default for \textit{nnU-Net}, as it also decreased performance. We note that spacing correction in \textit{nnU-Net} produces uniform spacing by inserting additional interpolated $z$-stacks.
As presented in Table \ref{tab:results}, the metrics obtained for the downscaled results are considerably better. 
We discuss the intuition for the improvements of the results in section~\ref{sec:summary}.
\begin{figure*}[h]
    \centering
    \includegraphics[width=0.9\linewidth]{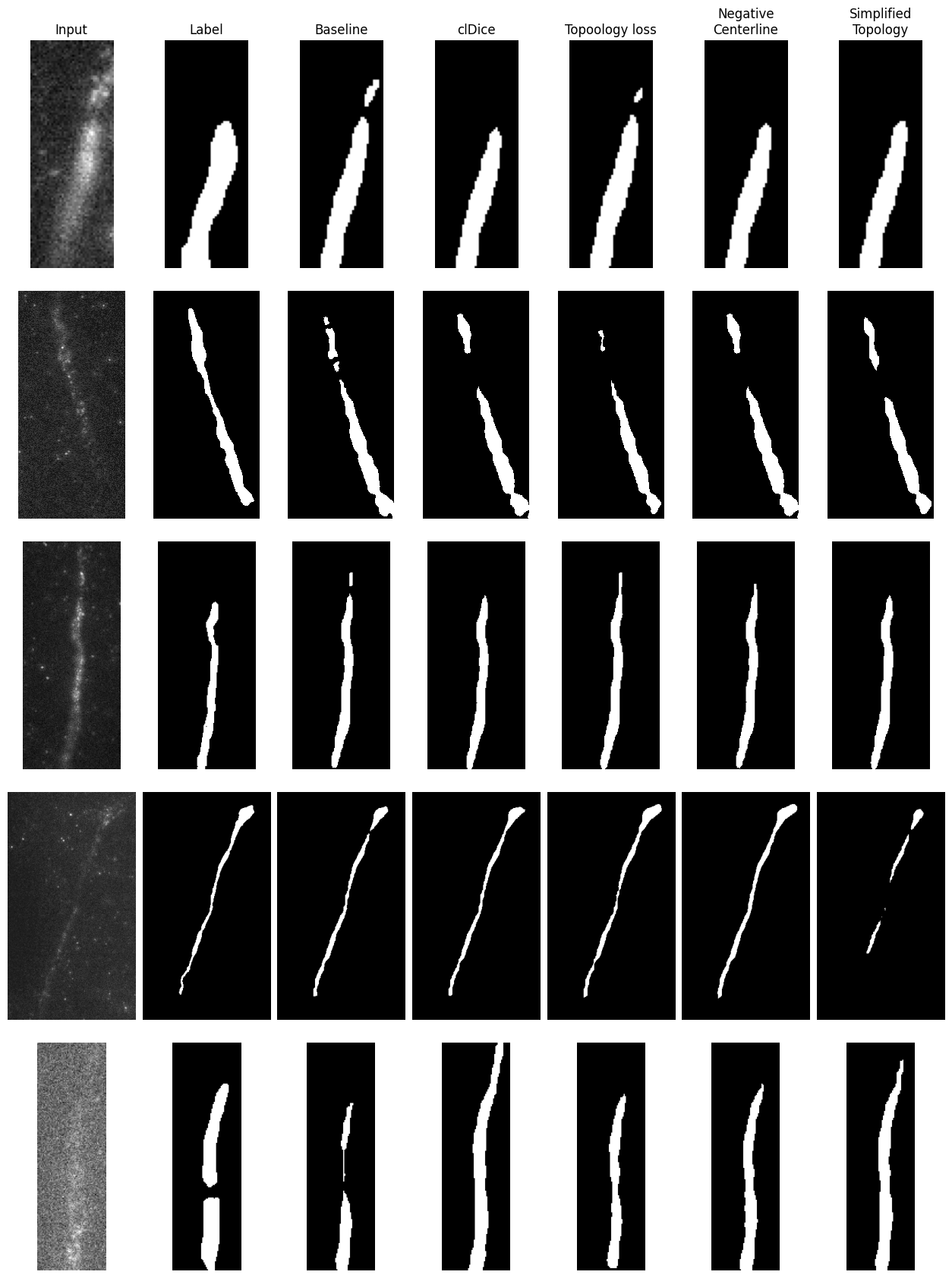}
    \caption{Comparison of the segmentations resulting from the proposed loss functions, displayed as a maximum projection from multiple $z$-stacks. Simple cases of distal discontinuities are often handled well. Proximal disconnectivity, as seen in the second row, is a more difficult task. Annotation of 3D volumes is a challenging procedure due to the two-dimensional nature of slice previews. As a result, in some cases, \eg row 3 and 5, some versions of the model outperform the ground truth label, by segmenting areas missed by humans. Such misalignments are anticipated and advocate for model improvement via active learning.}
    \label{fig:pred}
\end{figure*}
\begin{table*}[h]
\centering
    \caption{Comparison of the resulting metrics for each of the analysed loss functions. The result in bold highlights the best metric value for the given run type.}
    \label{tab:results}
\begin{tabular}{|c||c||c|c|c|c|c|}

  \hline
  Run type & Metric & \makecell{Baseline\\\textit{nnU-Net}} & clDice & \makecell{Topology\\loss} & \makecell{Negative\\Centerline Loss} & \makecell{Simplified\\Topology Loss} \\
  \hline \hline
  \multirow{6}{*}{3x downscaling} & Dice & \textbf{0.826} & 0.821 & 0.824 & 0.818 & 0.822  \\
  \cline{2-7}
  & \makecell{Wasserstein\\distance} & 55.5 & 54.0 & 54.4 & \textbf{42.9} & 49.3  \\  
  \cline{2-7}
  & \makecell{Strictly standardized\\mean difference} & 0.197 & 0.194 & 0.220 & \textbf{0.172} & 0.178 \\  
  \cline{2-7}
  & \makecell{Precision} & 0.822 & 0.830 & 0.844 & \textbf{0.862} & 0.857 \\  
  \cline{2-7}
  & \makecell{Recall} & 0.892 & \textbf{0.902} & 0.896 & 0.879 & 0.888 \\  
  \cline{2-7}
  & \makecell{Overlapping instances} & \textbf{1.038} & 1.049 & 1.045 & 1.042 & 1.054 \\  
  \hline

\multirow{6}{*}{No downscaling} & Dice & 0.799 & 0.800 & 0.803 & \textbf{0.803} & 0.802  \\
  \cline{2-7}
  & \makecell{Wasserstein\\distance} & 141.44 & 138.75 & 146.84 & \textbf{122.24} & 138.53  \\  
  \cline{2-7}
  & \makecell{Strictly standardized\\mean difference} & 0.517 & 0.496 & 0.526 & \textbf{0.448} & 0.494 \\  
  \cline{2-7}
    & \makecell{Precision} & 0.586 & 0.609 & 0.593 & \textbf{0.634} & 0.590 \\  
      \cline{2-7}

  & \makecell{Recall} & 0.898 & 0.837 & \textbf{0.899} & 0.897 & 0.891 \\  
  \cline{2-7}
  & \makecell{Overlapping instances} & 1.152 & 1.131 & 1.121 & \textbf{1.107} & 1.114 \\  
  \hline

\multirow{6}{*}{\makecell{No downscaling\\Small dataset}} & Dice & 0.633 & \textbf{0.670} & 0.627 & 0.612 & 0.658  \\
  \cline{2-7}
  & \makecell{Wasserstein\\distance} & 244.33 & 224.29 & 237.47 & 207.17 & \textbf{167.40}  \\  
  \cline{2-7}
  & \makecell{Strictly standardized\\mean difference} & 0.980 & 0.872 & 0.959 & 0.684 & \textbf{0.632} \\  
  \cline{2-7}

    & \makecell{Precision} & 0.384 & 0.437 & 0.392 & 0.356 & \textbf{0.563} \\  
  \cline{2-7}
    & \makecell{Recall} & \textbf{0.906} & 0.879 & 0.890 & 0.793 & 0.858 \\  
  \cline{2-7}

  & \makecell{Overlapping instances} & 1.520 & 1.382 & 1.501 & \textbf{1.075} & 1.248 \\  
  \hline
\end{tabular}
\end{table*}
\subsection{General results}
Once again referring to Table \ref{tab:results} we see that the application of the \textit{Negative Centerline Loss} yields better results across both scales, resulting in the lowest Wasserstein distance and highest precision. Likewise, the \textit{Simplified Topology Loss} also improves the model's output with respect to other methods. Importantly, this improvement generally does not come with the cost of considerably lowering the Dice score. 
Except for the Wasserstein distance, which directly corresponds to the quality of the output and its suitability for downstream research questions, like the comparison of AIS length distributions, the most important metrics for continuity are precision and the number of overlapping instances. In this case, we see that most prediction instances are matched with some label instances, while preserving high recall and Dice scores. Since instances can only be matched once, higher precision generally means that fewer instances are split. This is further confirmed by the average number of overlapping instances. Though we acknowledge that the baseline model also performs well in the downscaled case, the lower Wasserstein distance and strictly standardized mean difference make both the \textit{Simplified Topology Loss} and \textit{Negative Centerline Loss} versions preferable for downstream applications.

Most discontinuities can be classified into two categories, discontinuities at the ends of AIS, where a small region is detached, which degrades the length measurement to a small degree, and discontinuities in the middle of the AIS, which is much more severe. Both loss functions are able to consistently alleviate the first type of discontinuity, but severe cases of discontinuities in the middle of the AIS are more difficult to remove. One such example of discontinuity can be seen in row 2 of Figure \ref{fig:pred}, where both loss methods solve the discontinuity problem in a smaller, separated part of the AIS, where the discontinuities are separated by smaller gaps, but still struggle to connect the instances that are further away.
We also show an additional run in which the size of the dataset used for training and validation was halved, resulting in just four training volumes per fold. In this case, we see that the \textit{Simplified Topology Loss} outperforms other methods by a large margin. In biomedical applications, the datasets are often small or the labels scarce. We argue that the \textit{Simplified Topology Loss} is better suited for runs without much available data; the two loss functions combined can be used in active learning scenarios in which the model is iteratively improved by manually correcting its predictions. In those scenarios, we recommend using the \textit{Simplified Topology Loss} for initial training efforts, and later switching to the \textit{Negative Centerline Loss} for training when more data becomes available. 

\subsection{Remarks on performance}
The baseline \textit{nnU-Net} runtime required on average 150 seconds per epoch. \textit{Negative Centerline Loss}, \textit{clDice} and \textit{Simplified Topology Loss} all required roughly 13\% more time per epoch. Though the asymptotic guarantees of the \textit{Simplified Topology Loss} are better than those of the centerline-based methods (in most cases the inverse Ackermann function remains smaller than the radius of the objects), its runtime is comparable, since the centerline-based methods benefit from GPU acceleration. 
\section{Summary and future work}
\label{sec:summary}
In this work, we addressed one of the critical issues with segmenting elongated structures, which is instance continuity. 
We proposed two loss functions, \textit{Simplified Topology Loss} and \textit{Negative Centerline Loss}, both of which improve connectivity and length alignment on axon initial segment light-sheet microscopy data. Both losses can be straightforwardly implemented with available machine learning frameworks, are agnostic to imaging modality and biological context, and can therefore be applied to other tasks involving elongated structures. This makes them potentially applicable to other domains, such as vascular or road network segmentation, where continuity is essential and interruptions in signal are common. We recommend incorporating the \textit{Negative Centerline Loss} in any scenario that requires continuity preservation, and suggest the \textit{Simplified Topology Loss} if the dataset is small. 

This study answers a certain need for continuous segmentation, but also poses multiple questions for further research. Namely, we aim to follow this study with a broader evaluation of the effect of downscaling and spacing correction on segmentation quality. We note that 3D CNN models, despite being larger in parameter count, are usually shallower, and employ fewer filters than their 2D counterparts, due to the memory limitations and smaller datasets. We surmise that this might explain why directing the model to train strictly on the disconnected regions improves performance; whereas overparametrized 2D models usually have the capacity to learn both valid segmentation and preserve continuity when signal is lacking, in 3D scenarios this task is more difficult. This conjecture remains to be studied in future works. 

\subsection{Acknowledgements}
We publish the source code used for the experiments as a fork of the \textit{nnU-Net} repository. The codebase can be accessed on GitHub: \href{http://github.com/HelmholtzAI-Consultants-Munich/nnUNet-continuity-loss}{github.com/HelmholtzAI-Consultants-Munich/nnUNet-continuity-loss} and includes all components used in this study, including the implementation of \textit{clDice} and topology losses. This work was supported by the European Research Council (AXPLAST 803870 to JG), the Deutsche Forschungsgemeinschaft (SFB1089, SPP2411 to JG) and the Land Nordrhein-Westfalen (iBehave to JG). We are grateful to Dr. Bastian Grossenbacher-Rieck for valuable insights on topological machine learning and Stefan Fischer for sharing his knowledge of the \textit{nnU-Net} framework.

Training convolutional models with 3D data is a computationally intensive task. This study was planned to limit the number of runs while maintaining accurate reporting. We estimate that the total energy required to perform this study was approximately 2500 kWh, emitting approximately 875 kg of CO\textsubscript{2}.

\clearpage
\raggedbottom
{
    \small
    \bibliographystyle{ieeenat_fullname}
    \bibliography{main}

}

\end{document}